\documentclass[10pt,twocolumn,letterpaper]{article}

\usepackage{cvpr}              

\usepackage{booktabs}

\usepackage{xcolor,colortbl}
\usepackage{multirow}
\usepackage{tabularx}
\usepackage{array}
\usepackage{amsmath}
\usepackage{makecell}
\usepackage{float}
\usepackage{diagbox}
\usepackage{lipsum}
\definecolor{lightergray}{cmyk}{0,0.0,0.0,0.14}

\makeatletter
\renewcommand{\paragraph}{%
  \@startsection{paragraph}{4}%
  {\z@}{0.5ex \@plus 0.5ex \@minus .2ex}{-1em}%
  {\normalfont\normalsize\bfseries}%
}
\makeatother

\definecolor{cvprblue}{rgb}{0.21,0.49,0.74}
\usepackage[pagebackref,breaklinks,colorlinks,allcolors=cvprblue]{hyperref}

\def\paperID{} 
\def\confName{CVPR}
\def\confYear{2026}

\title{CPUBone: Efficient Vision Backbone Design for Devices with Low Parallelization Capabilities}

\author{Moritz Nottebaum\textsuperscript{1}\\
{\tt\small nottebaum.moritz@spes.uniud.it}
\and Matteo Dunnhofer\textsuperscript{1,2}\\
{\tt\small matteo.dunnhofer@uniud.it}
\and Christian Micheloni\textsuperscript{1}\\
{\tt\small christian.micheloni@uniud.it}
\\
\textsuperscript{1}University of Udine, Italy \quad
\textsuperscript{2}York University, Canada
}

\begin{document}
\maketitle


\begin{abstract}
  Recent research on vision backbone architectures has predominantly focused on optimizing efficiency for hardware platforms with high parallel processing capabilities. This category increasingly includes embedded systems such as mobile phones and embedded AI accelerator modules.
In contrast, CPUs do not have the possibility to parallelize operations in the same manner, wherefore models benefit from a specific design philosophy that balances amount of operations (MACs) and hardware-efficient execution by having high MACs per second (MACpS).
In pursuit of this, we investigate two modifications to standard convolutions, aimed at reducing computational cost: grouping convolutions and reducing kernel sizes.
While both adaptations substantially decrease the total number of MACs required for inference, sustaining low latency necessitates preserving hardware-efficiency.
Our experiments across diverse CPU devices confirm that these adaptations successfully retain high hardware-efficiency on CPUs.
Based on these insights, we introduce CPUBone, a new family of vision backbone models optimized for CPU-based inference.
CPUBone achieves state-of-the-art Speed-Accuracy Trade-offs (SATs) across a wide range of CPU devices and effectively transfers its efficiency to downstream tasks such as object detection and semantic segmentation.
Models and code are available at \url{https://github.com/altair199797/CPUBone}.
\end{abstract}

\begin{figure}
\centering
\includegraphics[width=\linewidth]{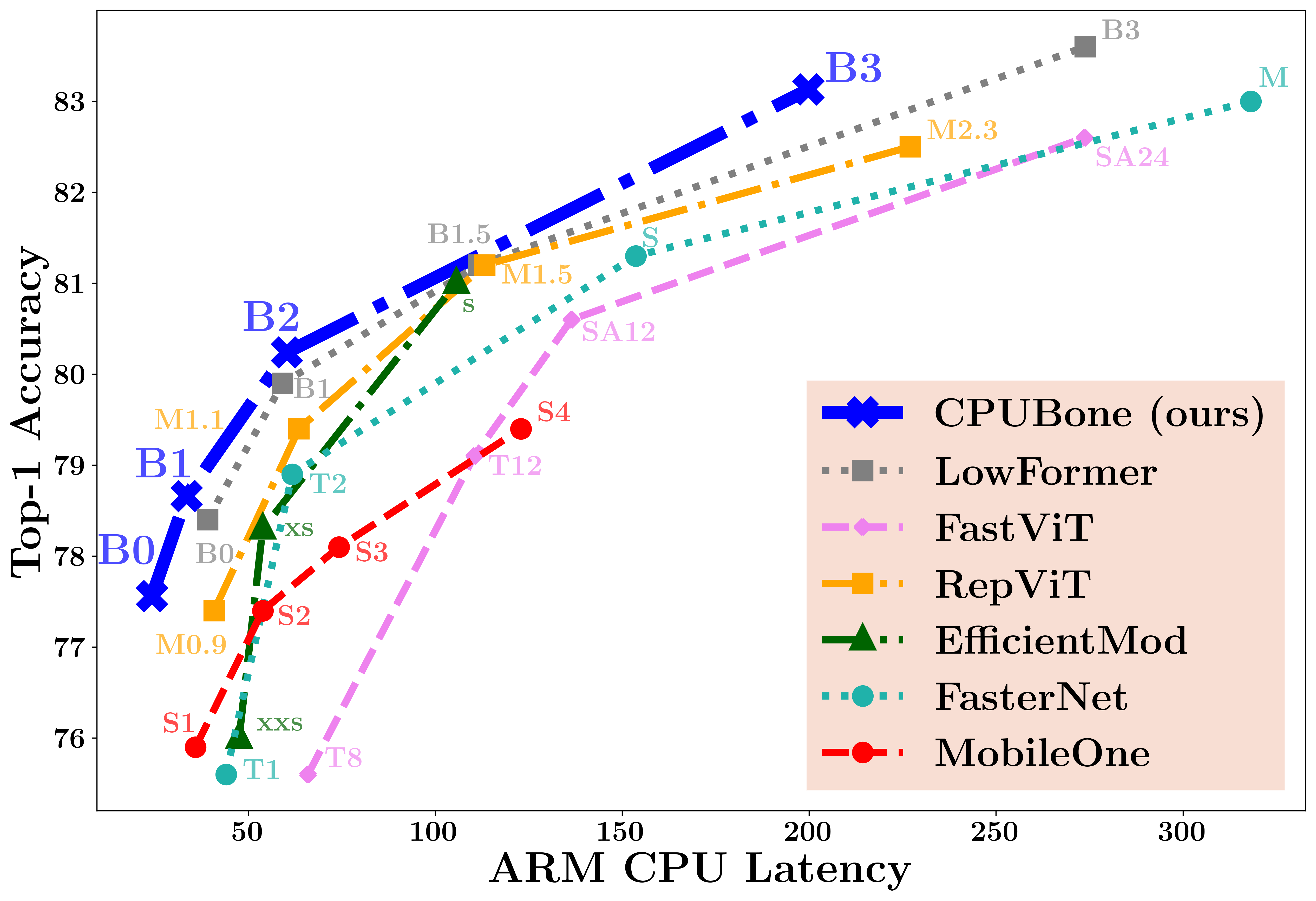}
\caption{ Comparison of ARM CPU latency and ImageNet top-1 accuracy of recent image classification architectures, including our proposed CPUBone models. Markers refer to different model sizes within each architecture family. CPUBone consistently achieves lower latency at comparable accuracy, highlighting its CPU-optimized design.  }
\label{fig:intro_figure}
\vspace{-0.2cm}
\end{figure}

\section{Introduction}
\label{sec:intro}


The design of vision backbone architectures has advanced significantly in recent years \cite{efficientvit, CvT,repvit,efficientnetv2,fasternet}, with increasing emphasis on improving the Speed-Accuracy Trade-off (SAT) of models.
However, numerous publications have shown that relying solely on the number of multiply and accumulate (MAC) operations is insufficient to assess model efficiency \cite{mobilenetv4, nottebaum2025lowformer, mobileone}, as execution performance varies substantially across hardware architectures due to differences in design and computational characteristics \cite{mobilenetv4, nottebaum2025lowformer}.
As a result, recent vision backbone architectures often target optimization for specific device classes, tailoring their designs to the characteristics and constraints of particular hardware subsets \cite{repvit, fastvit,mobileformer, efficientmodulation}.

However, the majority of these backbone architectures are still designed to exploit high degrees of computational parallelism to enhance runtime efficiency \cite{repvit, fastvit, mobileone}.
This trend is driven by the fact that, beyond Graphics Processing Units (GPUs), embedded systems such as mobile phones and AI modules like NVIDIA Jetson devices have increasingly demonstrated substantial parallel processing capabilities -- often matching or exceeding traditional GPUs in inference latency for various vision backbone architectures \cite{fastvit}.



On the other side, Central Processing Units (CPUs) -- which remain widely used \cite{smattracker,yolov10} in many practical and resource-constrained  scenarios  -- operate under fundamentally different architectural constraints \cite{fasternet}. In contrast to GPUs, CPUs offer limited parallelism, wherefore existing models optimized for devices with high parallelization capabilities \cite{fastvit, mobileone, fasternet} often fail to translate their efficiency to CPU-based systems \cite{efficientmodulation}. This is largely due to the excessive number of MAC operations, which become a performance bottleneck under the CPU's limited concurrency \cite{fasternet}.

To address this gap, we propose CPUBone (Section \ref{sec:cpubone}), a novel architecture family of vision backbone models specifically designed to optimize performance on CPUs by minimizing MAC count, while maintaining a hardware-efficient execution of the MAC operations.
Following \cite{fasternet}, we measure hardware-efficiency for execution time by counting how many MACs are processed per second and will further abbreviate this metric by MACpS (\textbf{MAC}s \textbf{p}er \textbf{s}econd).


A key component  of CPUBone is a modification of the mobile inverted bottleneck block (MBConv) \cite{mobilenetv2}, whose efficiency benefits on CPU-based systems are demonstrated in Section \ref{sec:mbconvblock}.
CPUBone models consistently achieve state-of-the-art results in the SAT across a variety of CPU devices and maintain their effectiveness when applied to downstream tasks such as object detection and semantic segmentation.

Our contributions can be summarized as follows:

\begin{itemize}

    \item We introduce two new MBConv variants that achieve both reduced MAC count  and hardware-efficient execution (high MACpS): Grouped Fused MBConv (GrFuMBConv) block and  Grouped MBConv (GrMBConv) block. 
    \item We conduct a comprehensive analysis of hardware-efficiency (MACpS) on both CPUs and GPU for the proposed MBConv variants, and further investigate the effect of reducing convolutional kernel sizes from $3\times3$ to $2\times2$.
    \item We present the vision backbone family CPUBone, which achieves state-of-the-art SATs on CPU-based devices.

\end{itemize}

\section{Related Work}

\subsection{Hardware-aware Architecture Design}

Recent studies have shifted from using theoretical operation counts (e.g., MACs) towards direct measurements of latency and throughput to evaluate execution efficiency \cite{mobileone,efficientvit, edgevit}. This transition reflects the growing recognition that a model’s efficiency can vary substantially across hardware platforms, as hardware characteristics strongly influence execution behaviour \cite{nottebaum2025lowformer, mobilenetv4}. Consequently, many works have explored hardware-aware model design, adapting architectures to specific devices or classes of hardware to maximize practical performance \cite{fasternet,repvit}.


While models such as FasterNet \cite{fasternet} and EfficientMod \cite{efficientmodulation} achieve competitive performance across multiple hardware categories -- particularly GPU throughput/latency and CPU latency --   there has been an increasing trend towards optimizing architectures primarily for mobile devices \cite{mobilenetv4, mobileone, repvit}. For instance, MobileNetV4 \cite{mobilenetv4} evaluates performance across a wide range of mobile devices, whereas RepViT \cite{repvit}, next to reporting GPU throughput, focuses on device-specific latency evaluation on the iPhone 12. Other architectures, such as SHViT \cite{shvit}, concentrate on throughput optimization, however on CPU and GPU devices. On the other side MobileOne \cite{mobileone} demonstrates that hardware-aware design can simultaneously benefit mobile phones, CPUs, and GPUs, highlighting the potential of flexible, multi-platform efficient architectures. \newline
We, however, focus our study specifically on CPU devices, which are inherently limited in their parallelization capabilities \cite{lopez2013toward} and differ substantially in that regard from GPUs and embedded AI accelerators (e.g. Nvidia Jetson devices), that feature significantly more compute cores. Additionally, unlike many mobile chips, CPUs usually lack  an integrated neural processing unit (NPU) and are unable to match the low-latency performance of modern mobile devices \cite{shvit}.
Overall, these factors make hardware-aware architecture design uniquely challenging on CPUs.




\subsection{Efficient Micro Design}
Many state-of-the-art vision backbone architectures adopt relatively simple macro designs while relying on sophisticated micro-architectural blocks to achieve high performance  \cite{fasternet, repvit, ghostnetv2, mobilenetv2, resnetpaper}.
Among the most widely used are the MBConv block \cite{mobilenetv2} and the ResNet bottleneck block \cite{resnetpaper}. 
Recently, \citeauthor{fasternet} \cite{fasternet} introduced the Partial Convolution (PConv), which has a reduced  MAC count compared to a standard convolution, as it only convolves over a portion of the channels, and is followed by lightweight pointwise convolutions. Next to a low MAC count, they demonstrate efficient execution with high MACpS on GPU, Intel CPU and ARM CPU platforms.
On the other side, the RepViT-block \cite{repvit} adopts the MetaFormer structure \cite{metaformer}, with a lightweight depthwise convolution as token mixer, achieving state-of-the-art results on the iPhone 12.

Instead of using depthwise convolutions (RepViT \cite{repvit}) or processing only a portion of the channels (PConv \cite{fasternet}), we reduce the MAC count of convolutions by increasing the groups parameter and reducing the kernel size. Similarly to \cite{fasternet}, we analyze how these adaptations influence MACpS on CPU and GPU devices.

\begin{figure}[ht]
\centering
\includegraphics[width=6cm]{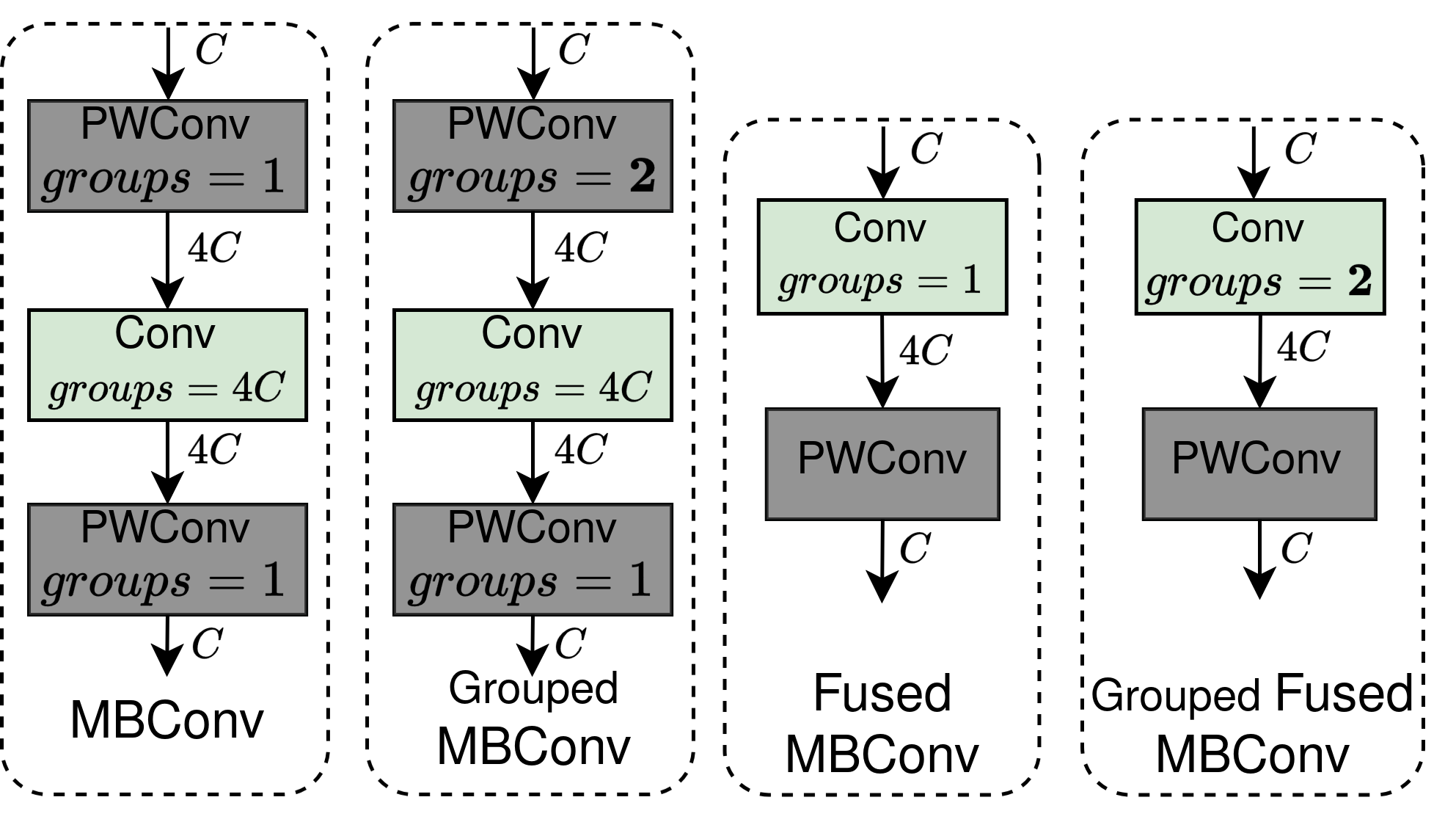}
\caption{ Design of MBConv variants, including the two proposed Grouped MBConv (GrMBConv) and Grouped Fused MBConv (GrFuMBConv). In both grouped variants, the first convolution is configured with $groups=2$. All variants expand the channel dimension in the first convolution by the expansion factor (set to four in this  figure) and reduce it again by the expansion factor in the last convolution.}
\label{fig:mbconvs_design}
\vspace{-0.4cm}
\end{figure}

\section{CPU-Efficient Design Strategies}
\label{sec:mbconvblock}

A high MAC count can significantly hinder fast execution on devices with limited parallelization capabilities, such as CPUs.
To overcome this we propose two adaptations to the standard configuration of convolutions: setting $groups=2$ and reducing the kernel size from the common $3\times3$ kernel \cite{CvT} to a $2\times2$ kernel.
\paragraph{Overview of Grouped Convolutions.}
The number of $groups$ in a convolutional layers is a decisive parameter, as it has a strong influence on the MAC count of the convolution. When $groups=1$, every output feature map is a combination of all input feature maps, while when $groups=C_{in}$ (where $C_{in}$ is the number of input channels), the convolutions becomes depthwise (DWConv). Depthwise convolutions process each input channel independently to produce the corresponding output channel. On the other side, for values between $1$ and $C_{in}$, the convolution is divided into $groups$ separate ungrouped sub-convolutions, where each sub-convolution processes a disjoint subset of the channels of size $C_{in}/groups$. Input channel dimension and output channel dimension always must both be divisible by $groups$.

In the following sub-sections, we first quantify how grouping and a smaller kernel size reduce  the MAC count compared to a standard ungrouped $3\times3$ convolution (see Subsection \ref{subsec:mac_calculation}).
We then analyze whether these reductions affect hardware-efficiency -- specifically, how they affect MACpS (see Section \ref{subsec:hardwareeff_all}).

\paragraph{Grouping for MBConv Block.}
In order to narrow down the configuration space we specifically analyze how the hardware-efficiency changes, when setting $groups=2$ in the first convolution of the mobile inverted bottleneck (MBConv) block \cite{mobilenetv2}  and the fused mobile inverted bottleneck (FuMBConv)  block \cite{fusedmbconv}. We chose both blocks, as they are popular components in vision backbones, due to their high efficiency \cite{efficientvit, nottebaum2025lowformer, efficientnetv2}.
The grouped variants are denoted as Grouped MBConv (GrMBConv) and Grouped Fused MBConv (GrFuMBConv), as illustrated in Figure \ref{fig:mbconvs_design}.



\subsection{Effect of Groups and Kernel Size on MAC Count}
\label{subsec:mac_calculation}
\paragraph{Grouping.}
The MAC  count ($M$) of a single convolution with no bias is given by
\begin{equation}\label{eq:maccount_general}
\begin{aligned}
   M =& \frac{K_{H} \times K_W \times C_{in} \times C_{out} \times  H_{out} \times W_{out} }{groups} ,  \\
\end{aligned}
\end{equation}
where $C_{out}$ is the output channel dimension, $K_H$ the height of the kernel and $K_W$ the width of the kernel.
By treating all variables except $groups$ in Equation~\eqref{eq:maccount_general} as constants, we can transform it and get a clearer understanding how the $groups$ parameter influences MAC count:

\begin{equation}\label{eq:maccount_grouping}
\begin{aligned}
   M \propto& \frac{1}{groups} .  \\
\end{aligned}
\end{equation}
Equation~\eqref{eq:maccount_grouping} shows that the MAC count ($M$) is inversely proportional to the number of groups. Therefore by increasing the number of groups to two, we halve the number of MAC operations required for execution. We can now apply Equation~\eqref{eq:maccount_general}
 to calculate how much less MACs the grouped MBConv variants have, compared to their ungrouped counterpart, leaving aside normalization and activation functions (see supplementary material for exact calculation). The GrFuMBConv block has 45\% less MACs than the FuMBConv, independent of the channel dimension. The GrMBConv has 23\% less MACs than the MBConv block, averaged over the five channel dimensions featured in Table \ref{tab:mbconv_eff_exp}.

\paragraph{Kernel Size.}

By treating all variables in Equation~\eqref{eq:maccount_general} unrelated to the kernel size as constants, we obtain:

\begin{equation}\label{eq:maccount_kernel}
\begin{aligned}
M &\propto K_{H} \times K_{W}.
\end{aligned}
\end{equation}
From this we can see, that the MAC count of a convolution is also proportional to the product of the kernel dimensions.
Consequently, reducing the kernel size directly lowers computational cost; for instance replacing a $3\times3$ kernel ($K^2=9$) with a $2\times2$ kernel ($K^2=4$) decreases the MAC count of the convolution by approximately $(1- 4/9) \sim 56\%$.
Similarly to grouping, we also apply the reduced  $2\times2$ kernel to the FuMBConv and GrFuMBConv block. While the total MAC of the FuMBConv is reduced by exactly 50\%, the GrFuMBConv block has approximately 46\% less MACs, compared to a $3\times3$ kernel (see supplementary material for exact calculation).


\begin{table*}[ht]
\centering
\small
\setlength{\tabcolsep}{6pt}
\renewcommand{\arraystretch}{1.25}

\begin{tabular}{c|l|ccccc|c}
\Xhline{1.5pt}
\multicolumn{2}{c|}{\textbf{Grouping Experiment}} & \multicolumn{5}{c|}{\textbf{Channel Dimension}} & \\
\cline{3-7}

\multicolumn{1}{c}{\textbf{Device}} & \multicolumn{1}{c|}{\textbf{Variant}} & 32 & 64 & 128 & 256 & 512 & \textbf{Avg.} \\
\Xhline{1.5pt}

    & \textit{\textcolor{gray}{MBConv}} & \textbf{7.8} & \textbf{13.3} & \textbf{21.4} & \textbf{26.4} & \textbf{26.2} & 19.0 \\[-2pt]
  Pi5 \textbf{CPU}  & \textit{\textbf{Gr}\textcolor{gray}{MBConv}} & 6.3 & 10.6 & 16.7 & 22.1 & 23.3 & 15.8 \footnotesize \textbf{(-16\%)} \\ 
    \cline{2-8}
 (MMACs/ms)   & \textit{\textbf{Fu}\textcolor{gray}{MBConv}} & \textbf{40.6} & \textbf{45.3} & 40.7 & 32.4 & 24.5 & 36.7 \\[-2pt]
    & \textit{\textbf{GrFu}\textcolor{gray}{MBConv}} & 31.0 & 42.1 & \textbf{44.4} & \textbf{34.7} & \textbf{30.4} & 36.5 \footnotesize \textbf{(-0\%)} \\
\Xhline{1.2pt}

    & \textit{\textcolor{gray}{MBConv}} & \textbf{51.0} & \textbf{59.8} & \textbf{60.9} & 53.4 & 44.5 & 53.9 \\[-2pt]
   Pixel 4 \textbf{CPU} & \textit{\textbf{Gr}\textcolor{gray}{MBConv}} & 45.3 & 56.7 & 59.2 & \textbf{58.3} & \textbf{45.9} & 53.1 \footnotesize \textbf{(-1\%)} \\
\noalign{\setlength{\arrayrulewidth}{0.1pt}}
\cline{2-8}
\noalign{\setlength{\arrayrulewidth}{0.4pt}}
   (MMACs/ms) & \textit{\textbf{Fu}\textcolor{gray}{MBConv}} & \textbf{67.3} & 63.0 & 53.2 & 39.9 & 34.1 & 51.5 \\[-2pt]
    & \textit{\textbf{GrFu}\textcolor{gray}{MBConv}} & 64.3 & \textbf{65.6} & \textbf{61.6} & \textbf{43.2} & \textbf{36.2} & 54.2 \footnotesize \textbf{(+5\%)} \\
\Xhline{1.2pt}

    & \textit{\textcolor{gray}{MBConv}} & \textbf{78.0} & \textbf{111.6} & \textbf{124.5} & \textbf{129.0} & \textbf{126.4} & 113.9 \\[-2pt]
  Snapdragon 8 \textbf{CPU}  & \textit{\textbf{Gr}\textcolor{gray}{MBConv}} & 66.3 & 105.9 & 120.7 & 125.0 & 125.3 & 108.6 \footnotesize \textbf{(-5\%)} \\
\noalign{\setlength{\arrayrulewidth}{0.1pt}}
\cline{2-8}
\noalign{\setlength{\arrayrulewidth}{0.4pt}}
  (MMACs/ms)  & \textit{\textbf{Fu}\textcolor{gray}{MBConv}} & \textbf{121.4} & 128.8 & 133.4 & \textbf{130.3} & \textbf{115.4} & 125.9 \\[-2pt]
    & \textit{\textbf{GrFu}\textcolor{gray}{MBConv}} & 114.0 & \textbf{129.4} & \textbf{133.8} & 125.1 & 113.9 & 123.2 \footnotesize \textbf{(-2\%)} \\
\Xhline{1.2pt}

    & \textit{\textcolor{gray}{MBConv}} & \textbf{34.2} & \textbf{112.3} & \textbf{428.6} & \textbf{1747.3} & \textbf{3494.0} & 1163.3 \\[-2pt]
 TITAN RTX \textbf{GPU}   & \textit{\textbf{Gr}\textcolor{gray}{MBConv}} & 25.0 & 88.4 & 308.8 & 1260.6 & 2679.1 & 872.4 \footnotesize \textbf{(-25\%)} \\
\noalign{\setlength{\arrayrulewidth}{0.1pt}}
\cline{2-8}
\noalign{\setlength{\arrayrulewidth}{0.4pt}}
  (MMACs/ms)  & \textit{\textbf{Fu}\textcolor{gray}{MBConv}} & \textbf{185.1} & \textbf{622.6} & \textbf{2394.6} & \textbf{3612.2} & 3849.0 & 2132.7 \\[-2pt]
    & \textit{\textbf{GrFu}\textcolor{gray}{MBConv}} & 96.8 & 330.3 & 1307.3 & 2963.3 & \textbf{3878.9} & 1715.3 \footnotesize \textbf{(-20\%)} \\
\Xhline{1.5pt}
\end{tabular}

\caption{Execution efficiency of MACs, measured in MMACs/ms, for the four MBConv variants (MBConv, GrMBConv, FuMBConv and GrFuMBConv) for different channel dimensions. Bold entries indicate whether the grouped or ungrouped variant has higher MACpS.}
\label{tab:mbconv_eff_exp}
\vspace{-0.2cm}
\end{table*}

\subsection{Hardware-Efficiency Analysis}
\label{subsec:hardwareeff_all}
While grouping and a smaller kernel size significantly reduce the MAC count of a convolution, it is not clear, if this advantage also translates to a significant reduction of latency. This is due to factors like degree of parallelism and memory access costs \cite{nottebaum2025lowformer, mobilenetv4, fasternet}. This phenomena can for example be observed for DWConvs \cite{nottebaum2025lowformer}, which have considerably lower MACpS compared to ungrouped convolutions ($groups=1$). Therefore, it is crucial to analyze how grouping or smaller kernel sizes affect MACpS.

\paragraph{Execution Time Measurement.}
In order to quantify hardware-efficiency in execution (MACpS), we need to measure latency. We feature several CPU devices for this:
 the CPU of the Raspberry Pi 5, the CPU of the Google Pixel 4, and the CPU of the Snapdragon 8 Elite QRD. We further compare these results with latency measurements on a device with high parallelization capability -- the Nvidia TITAN RTX GPU.
 Latency is always measured with a batch size of 1.


\subsubsection{Effect of Grouping on Hardware-Efficiency}
\label{subsubsec:grouping_hardwareeff}
\paragraph{Setting.}
In Table \ref{tab:mbconv_eff_exp}, we compare the MACpS of the MBConv and FuMBConv block with their grouped counterparts (GrMBConv and GrFuMBConv). We do so by measuring the number of MMACs (million MACs) each architectural block executes per millisecond (MMAC/ms).
While we fix the operating resolution  to $14\times14$, we feature five different input channel dimensions (32, 64, 128, 256 and 512) and an expansion factor of 4 (see Figure \ref{fig:mbconvs_design}).

\paragraph{Observation 1.}
Across all CPU devices and channel dimensions, GrMBConv have on average 5\% lower MACpS compared to MBConv, while having 23\% less MACs.
On the other side, GrFuMBConv have 1\% higher MACpS than FuMBConv, while having 45\% less MACs.

\paragraph{Observation 2.}
Averaged across all CPU devices, the fused variants (FuMBConv and GrFuMBConv) have approximately  25\% lower MACpS, when the channel dimension is $\geq256$, compared to $<256$, while the unfused variants (MBConv and GrMBConv) have 27\% higher MACpS.
However, for channel dimensions below 256, the fused variants achieve  70\% higher MACpS than the unfused variants. This behaviour was previously observed by  \cite{nottebaum2025lowformer}, though only on devices with higher parallelization capabilities. 
\paragraph{Operating Resolution.}
We repeated the experiment for additional input resolutions ($7\times7$, $28\times28$ and $56\times56$) on ARM CPU (see supplementary material), yielding a similar result as in Table \ref{tab:mbconv_eff_exp}. Across the resolutions, GrMBConvs have 18\% lower MACpS than MBConv, while GrFuMBConvs have 2\% lower MACpS than FuMBConvs.

\paragraph{Effect on GPU.}
In Table \ref{tab:mbconv_eff_exp} we also feature the same experiment on GPU (Nvidia Titan RTX), however both observations do not hold on GPU anymore. The GrMBConv has on average 23\% less MACs, however on GPU it also has 25\% lower MACpS, making the GrMBConv block slower on average. Similarly, the GrFuMBConv has on average 20\% lower MACpS than the FuMBConv, making it still faster however due to having approximately half the MAC count. 
Additionally, similarly to what we observed on the CPU devices, the fused MBConv variants (GrFuMBConv \& FuMBConv) remain considerably more hardware-efficient (higher MACpS) than their unfused counterpart (GrMBConv \& MBConv), especially for channel dimensions below 256. They execute up to $5\times$ more MACs in the same time. We also repeated the same experiments for different resolutions (see supplementary material), leading to the same conclusion.

\paragraph{Higher Groups parameter.}
So far we have only considered $groups=2$, however we also repeated the experiment on an ARM CPU with $groups=4$ (see supplementary material). The averaged result we discussed with $groups=2$ are largely similar for $groups=4$, however the fused variants (GrFuMBConv \& FuMBConv) exhibit a different behaviour. For channel dimensions below 128,  GrFuMBConv has 20\% lower MACpS, when $groups=4$, compared to $groups=2$. However for channel dimensions over 128, they have 36\% higher MACpS.

\paragraph{Conclusion.}
The grouped variants achieve similar MACpS compared to their ungrouped counterparts (observation 1), while featuring a considerably lower MAC count, leading to improved execution time. However, on ARM CPU we observe, that the GrMBConv's hardware-efficiency degrades significantly (-16\% MACpS) compared to the MBConv block, making it nevertheless faster, as it has 23\% less MACs.
Additionally in order to optimize hardware efficiency, fused variants (FuMBConv and GrFuMBConv) are better suited for layers with fewer than 256 channels, while unfused variants (MBConv and GrMBConv) perform best with channel dimensions of 256 or more (observation 2).

\begin{table}[ht]
\centering
\setlength{\tabcolsep}{6.5pt}
\renewcommand{\arraystretch}{1.25}
\footnotesize
\begin{tabular}{cc|cccc|c}
    \Xhline{1.5pt}
    \multicolumn{2}{c|}{\textbf{Dwise}} & \multicolumn{4}{c|}{\textbf{Channel Dimension}} & \\[-3pt]
    \textbf{Resol.} & \textbf{Type} & 128 & 256 & 512 & 1024 & \textbf{Avg.} \\
    \Xhline{1.5pt}

    \multirow{2}{*}{\textbf{7×7}} 
        & \textit{\textcolor{gray}{nmk}} & \textbf{0.75} & \textbf{0.78} & \textbf{0.78} & \textbf{0.80} & \textbf{0.78} \\[-2pt]
        & \textit{\textcolor{gray}{smk}} & 0.66 & 0.70 & 0.70 & 0.73 & 0.70 \\
    \cline{1-7}

    \multirow{2}{*}{\textbf{14×14}} 
        & \textit{\textcolor{gray}{nmk}} & \textbf{1.26} & \textbf{1.28} &\textbf{ 1.27} & \textbf{1.24} & \textbf{1.26} \\[-2pt]
        & \textit{\textcolor{gray}{smk}} & 1.13 & 1.15 & 1.16 & 1.08 & 1.13 \\

      \cline{1-7}

    \multirow{2}{*}{\textbf{28×28}} 
        & \textit{\textcolor{gray}{nmk}} & \textbf{1.97} & \textbf{1.82} & \textbf{1.64} & \textbf{1.64} & \textbf{1.77} \\[-2pt]
        & \textit{\textcolor{gray}{smk}} & 1.73 & 1.50 & 1.26 & 1.19 & 1.42 \\

    \Xhline{1.5pt}
\end{tabular}

\caption{Execution efficiency experiment on ARM CPU on the effect of kernel size reduction for depthwise convolutions, by measuring MMACs/ms. \textbf{Resol.} refers to operating resolution, \textbf{Type} refers to the kernel size ($nmk=3\times3$, $smk=2\times2$). Bold entries indicate whether nmk or smk variant has higher MACpS.}
\label{tab:kernel_exp_depthwise}
\vspace{-0.2cm}
\end{table}

\begin{table}[ht]
\centering
\setlength{\tabcolsep}{5pt}
\renewcommand{\arraystretch}{1.25}
\footnotesize
\begin{tabular}{cl|cc|c!{\vrule width 1.5pt}cc|c}
\Xhline{1.5pt}
 & & \multicolumn{3}{c!{\vrule width 1.5pt}}{\textbf{Ungrouped}} & \multicolumn{3}{c}{\textbf{Groups=2}} \\
\cline{3-8}
\textbf{Resol.} & \textbf{Type} & \multicolumn{2}{c|}{\textbf{Channel Dim.}} &  & \multicolumn{2}{c|}{\textbf{Channel Dim.}} & \\
\cline{3-4} \cline{6-7}
 & & 128 & 256 &\textbf{Avg.} & 128 & 256 & \textbf{Avg.} \\
\Xhline{1.5pt}

\multirow{2}{*}{\textbf{7×7}}
 & \textit{\textcolor{gray}{nmk}} & 39.36 & 28.63 & 33.99 & \textbf{37.88} & 32.42 & 35.15 \\[-2pt]
 & \textit{\textcolor{gray}{smk}} & \textbf{41.21} & \textbf{33.13} & \textbf{37.17} & 36.11 & \textbf{36.42} & \textbf{36.27 }\\
\cline{1-8}

\multirow{2}{*}{\textbf{14×14}}
 & \textit{\textcolor{gray}{nmk}} & 40.90 & 33.45 & 37.17 & \textbf{44.56} & 35.21 & 39.88 \\[-2pt]
 & \textit{\textcolor{gray}{smk}} & \textbf{46.94} & \textbf{36.01} & \textbf{41.48} & 43.86 & \textbf{40.58} & \textbf{42.22} \\
\cline{1-8}

\multirow{2}{*}{\textbf{28×28}}
 & \textit{\textcolor{gray}{nmk}} & 40.70 & 29.23 & 34.96 & \textbf{47.45} & 36.37 & 41.91 \\[-2pt]
 & \textit{\textcolor{gray}{smk}} & \textbf{42.95} & \textbf{31.60} & \textbf{37.28} & 46.67 & \textbf{38.02} & \textbf{42.35} \\
\Xhline{1.5pt}
\end{tabular}

\caption{Execution efficiency experiment on ARM CPU on the effect of kernel size reduction. We test the GrFuMBConv (Groups=2) and FuMBConv (Ungrouped) block by measuring MMACs/ms.  \textbf{Resol.} refers to operating resolution, \textbf{Type} refers to the kernel size ($nmk=3\times3$, $smk=2\times2$). Bold entries indicate whether nmk or smk variant has higher MACpS.}
\label{tab:kernel_exp_groups2_fused}
\vspace{-0.2cm}
\end{table}


\subsection{Effect of Kernel Reduction on Hardware-Efficiency}
\label{subsubsec:kernel_hardwareeff}


\paragraph{Setting.}
As shown before Subsection \ref{subsec:mac_calculation}, reducing the kernel size considerably reduces the MAC count. However, similarly to grouping, we need to ensure that the MACpS do not deteriorate. 
In order to have a comprehensive analysis, also fitting to the previous grouping analysis and to our macro architecture, we feature three different kind of convolutions or blocks: Depthwise convolutions (as central part of the MBConv block, see Figure \ref{fig:mbconvs_design}), the FuMBConv block (see Figure \ref{fig:mbconvs_design}) and the GrFuMBConv block, with $groups=2$. Both MBConv versions feature an expansion factor of 4. 
For depthwise convolutions we span channel dimensions from 128 to 1024, because this is the usual range occurring in the later stages of vision backbones \cite{nottebaum2025lowformer, efficientvit, fastvit, shvit}. On the other side, for the FuMBConv and GrFuMBConv, we only test for channel dimensions 128 and 256, as our previous analysis in Subsection \ref{subsubsec:grouping_hardwareeff} as well as \cite{nottebaum2025lowformer} conclude that a channel dimension higher than 256 for fused MBConv versions results in a considerable reduction of MACpS, wherefore we deem it less relevant. 
In Table  \ref{tab:kernel_exp_depthwise} and \ref{tab:kernel_exp_groups2_fused} we compare two settings of kernel sizes: nmk, referring to $3\times3$ kernel and smk, referring to $2\times2$ kernel. All results are measured on the ARM CPU of the Raspberry Pi5, featuring a batch size of 1.

\paragraph{Observation 1.}
While for the depthwise convolutions (see Table \ref{tab:kernel_exp_depthwise}) the reduction of the kernel size leads to a small reduction of MACpS (-10\%), for the FuMBConv and the GrFuMBConv ($groups=2$) block (see Table \ref{tab:kernel_exp_groups2_fused} ) the hardware-efficiency  increases on average. Since the kernel size reductions approximately halve the MAC count for the tested MBConv variants (as mentioned in Subsection \ref{subsec:mac_calculation}), they lead to a considerable latency improvement.

\paragraph{Observation 2.}
The absolute magnitude of MACpS for the depthwise convolution (Table \ref{tab:kernel_exp_depthwise}) is considerably lower than for the ungrouped or $groups=2$ convolutions (see Table \ref{tab:kernel_exp_groups2_fused}). The latter two have approximately $40\times$ higher MACpS, compared to the depthwise convolution. 
A similar gap can also be seen, when repeating the experiments on GPU (see supplementary material). Since depthwise convolutions exhibit low MACpS on devices with both low and with high parallelization capability, degree of parallelism is not the main factor, influencing the hardware-efficiency of depthwise convolutions. We believe memory access cost is mainly responsible for that. 

\begin{table*}[hbt!]

    \begin{center}
\resizebox{14cm}{!}{
    \begin{tabular}{c|cc|c|c|c|c}
        \Xhline{1.5pt}
         \multirow{2}{*}{Model} &  Params  & MACs & \multicolumn{1}{c|}{Pi5 CPU  }& \multicolumn{1}{|c|}{Pixel 7 Pro CPU}   & \multicolumn{1}{|c|}{Intel CPU }  & Top-1  \\
         & (M) & (M) & (ms) $\downarrow$  &  (ms) $\downarrow$ & (ms) $\downarrow$ &  (\%)  \\
        \Xhline{1.5pt}

        EffFormerV2-S0 \cite{efficientformerv2} & 3.5 & 400 & 36.3 &  39.4 &  7.9 &   73.7 \\
        
        GhostNetV2 x1.0 \cite{ghostnetv2} & 6.2 & 183 & 35.2 &  5.8   & 9.0   &  75.3 \\
        
        

        FastViT-T8 \cite{repvit} & 3.6 & 705 & 65.9 &  44.0  & 10.6  &  75.6 \\
        MobileViG-Ti \cite{mobilevig} & 5.3 & 661 & 48.2  & 45.2  &  9.2   &  75.7 \\
        EfficientMod-xxs \cite{efficientmodulation} & 4.7 & 579 & 47.5  & 25.3  & 9.2  &  76.0 \\

        FasterNet-T1 \cite{fasternet}  & 7.6 & 850 & 32.8 &  24.4  &  6.3   &  76.2 \\

        RepViT-M0.9 \cite{repvit} & 5.1 & 816 & 40.8 &  35.1  & 10.2  &  77.4 \\

       
       \rowcolor{lightergray}  CPUBone-B0 (ours) & 10.4 & 519 & 24.2 &  13.8 &  6.9 &   \textbf{77.6} \\

        \Xhline{1.5pt}

        EffFormerV2-S1 \cite{efficientformerv2} & 6.1 & 650 & 57.4  & 54.6   & 10.9  &  77.9 \\

        MobileViG-S \cite{mobilevig} & 7.3 & 983 & 73.9 &  69.5  & 13.0   &  78.2 \\

        EfficientMod-xs \cite{efficientmodulation} & 6.6 & 773 & 53.8  & 28.1  & 11.5  &  78.3 \\

        LowFormer-B0 \cite{nottebaum2025lowformer} & 14.1 & 944 & 39.1    & 20.8  & 9.4 &  78.4   \\

        \rowcolor{lightergray} CPUBone-B1 (ours) & 12.4 & 746 & 33.5   & 18.6 &  9.4   &  \textbf{78.7} \\

        \Xhline{1.5pt}
        FasterNet-T2 \cite{fasternet}  & 15.0 & 1910 & 61.7 &  28.5  &  10.1   &  78.9 \\
        
        MobileOne-S4 \cite{mobileone}  & 14.8 & 2978 & 122.9  & 43.6  &  13.3   &  79.4 \\

        RepViT-M1.1 \cite{repvit} & 8.2 & 1338 & 63.5  & 48.7 & 12.0  &  79.4 \\
        
         LowFormer-B1 \cite{nottebaum2025lowformer} & 17.9 & 1410 & 59.1  &  30.6  & 13.2 &  79.9   \\

           \rowcolor{lightergray} CPUBone-B2 (ours) & 23.9 & 1354 & 60.3   & 32.4   & 16.1 &   \textbf{80.3} \\
         \Xhline{1.5pt}   
        EffFormerV2-S2 \cite{efficientformerv2} & 12.6 & 1250 & 102.3 &  91.5   & 18.9   &  80.4 \\

        FastViT-SA12 \cite{fastvit} & 10.9 & 1943 & 136.4  & 86.2   & 20.0  &  80.6 \\


        LowFormer-B15 \cite{nottebaum2025lowformer} & 33.9 & 2573 & 111.6  &  56.8  & 22.8 &  81.2   \\
        FasterNet-S \cite{fasternet}  & 31.1 & 4560 & 153.6 &  70.7  &  20.6   &  81.3 \\
        
        BiFormer-T \cite{biformer} & 13.1 & 2200 &  523.9  &  578.6 &  178.0  & 81.4   \\
        
        Resnet101 \cite{resnetpaper} & 44.5 & 7801 &  293.8    & 115.6  & 30.8  & 81.9   \\

        RepViT-M2.3 \cite{repvit} & 22.9 & 4520 & 227.0  & 148.0   & 37.7  &  82.5 \\

        FasterNet-M \cite{fasternet}  & 53.5 & 4370 & 318.1  & 151.5   & 41.6  &  83.0 \\

         \rowcolor{lightergray} CPUBone-B3 (ours)  & 40.7 & 4054 & 199.8  & 83.1 &   34.1  &   \textbf{83.1} \\
        \Xhline{1.5pt}
        
        FasterNet-L \cite{fasternet}  & 93.5 & 7760 & 644.8 &  290.1  & 65.5  &  83.5 \\

        MIT-EfficientViT-B3 \cite{efficientvit} & 49.0 & 3953 & 340.2  & 98.2  & 44.1   &  83.5 \\

         LowFormer-B3 \cite{nottebaum2025lowformer} & 57.1 & 6098 & 273.8    & 124.0  & 44.4    & 83.6   \\
         BiFormer-S \cite{biformer} & 26.0 & 4500 &  1134.1    & 1290.0  & 391.0  & \textbf{83.8}   \\

        \Xhline{1.5pt}
    \end{tabular}}

    \end{center}
   
    \caption{Performance on ImageNet-1K validation set. Evaluation resolution is 224×224, except for FastViT models who operate on 256x256. Besides MobileVig \cite{mobilevig}, no distillation nor pretraining is used for fair comparison. Models are sorted and grouped by top-1 accuracy. The highest top-1 accuracy in each group is bold. }
    \label{tab:imagenetresults}
    
\end{table*}

\newpage
\paragraph{Effect on GPU.}
Similarly to how grouping reduces MACpS on GPU, reducing the kernel size deteriorates 
performance completely on GPU. To show that we repeat the same experiments of Tables \ref{tab:kernel_exp_depthwise} and \ref{tab:kernel_exp_groups2_fused} on the Nvidia TITAN RTX GPU (see supplementary material).
For depthwise convolutions, the MACpS reduces by 80\% approximately, making the smaller kernel version slower than the one with the original $3\times3$ kernel. Similarly, the FuMBConv and the GrFuMBConv block approximately have 50\% lower MACpS. Consequently, reducing the kernel size consistently leads to higher execution time on GPU, even though less MACs need to be executed.

\begin{figure*}[h]
\centering
\begin{minipage}[t]{0.55\textwidth}

  \centering
    \includegraphics[width=\textwidth]{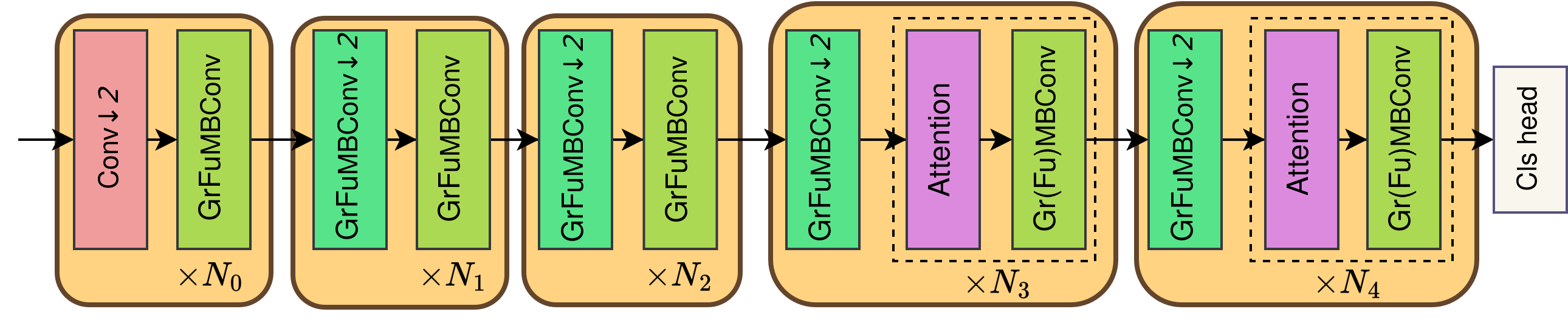}
    \captionof{figure}{CPUBone macro architecture design. Attention refers to LowFormer Attention \cite{nottebaum2025lowformer}.}
    \label{fig:architecture}

    \vspace{0.5em} 

    \setlength{\tabcolsep}{6pt}
    \renewcommand{\arraystretch}{1.25}
    \resizebox{!}{1.2cm}{
    \begin{tabular}{c|c|c|c|c}
        \Xhline{1.0pt}
         Stage  & B0  & B1 & B2 & B3  \\
         \Xhline{0.5pt}

        0 & C=16, N=0, \textbf{GrFu}  & C=16, N=0, \textbf{GrFu}  & C=20, N=0, \textbf{GrFu} & C=32, N=1, \textbf{GrFu} \\
            
        1 & C=32, N=0, \textbf{GrFu} & C=32, N=0, \textbf{GrFu} & C=40, N=0, \textbf{GrFu} & C=64, N=1, \textbf{GrFu}\\
        
        2 & C=64, N=0, \textbf{GrFu}  & C=64, N=0, \textbf{GrFu} & C=80, N=0, \textbf{GrFu} & C=128, N=2, \textbf{GrFu}\\
        
        3 & C=128, N=\textbf{3}, \textbf{GrFu} & C=128, N=\textbf{5}, \textbf{GrFu} & C=160, N=\textbf{6}, \textbf{GrFu} & C=256, N=\textbf{6}, \textbf{Gr}\\
        4 & C=256, N=\textbf{4}, \textbf{Gr} & C=256, N=\textbf{5}, \textbf{Gr} & C=320, N=\textbf{6}, \textbf{Gr} & C=512, N=\textbf{6}, \textbf{Gr}\\
        
         \Xhline{1.0pt}
    \end{tabular}}
    \captionof{table}{Specification of CPUBone architecture versions B0-B3. C, N, Gr and GrFu stand for channel dimension, the number of layers, GrMBConv and GrFuMBConv. Number of layer higher than two, are bold.}
    \label{tab:arch_specs}  
    
\end{minipage}
\hfill
\begin{minipage}[t]{0.4\textwidth}
\centering
    \resizebox{!}{2.9cm}{
    \begin{tabular}{c|c|c}
        \Xhline{1.5pt}
         \multirow{2}{*}{Backbone}  & \multicolumn{1}{c|}{Pi5 CPU Lat.} & mIoU    \\
         & (ms) &   (\%) \\
         \Xhline{1.0pt}

        ResNet50  \cite{resnetpaper} & 940.0   & 36.7 \\
        PVTv2-B0  \cite{pvtv2} & 587.7  & 37.2 \\
        \rowcolor{lightergray} CPUBone-B0 (ours) & \textbf{131.5}   &  \textbf{37.9} \\
        \Xhline{1.0pt}

        FastViT-SA12  \cite{fastvit} & 603.9   & 38.0 \\
        \rowcolor{lightergray} CPUBone-B1 (ours) & \textbf{189.9}   &  \textbf{39.2} \\
        \Xhline{1.0pt}


        RepViT-M1.1  \cite{repvit} & 468.9   & 40.6 \\
        FastViT-SA24  \cite{fastvit} & 1161.5   & 41.0 \\
        EdgeViT-XS \cite{edgevit}  & 461.5   & 41.4  \\
        FAT-B0 \cite{fat}  & 763.3   & 41.5 \\


        \rowcolor{lightergray} CPUBone-B2 (ours) & \textbf{338.2} &   \textbf{42.1} \\
        \Xhline{1.0pt}
        
        PVTv2-B1  \cite{pvtv2} & 1296.4   & 42.5 \\
        LowFormer-B2 \cite{nottebaum2025lowformer} & \textbf{808.1}  & 42.8 \\
      
        FAT-B1 \cite{fat}  & 1102.1  & 42.9 \\
        \rowcolor{lightergray} CPUBone-B3 (ours) & 1181.6  &  \textbf{44.1} \\
          \Xhline{1.5pt}

     
    \end{tabular}}

\caption{ Results on semantic segmentation, using Semantic FPN \cite{semanticfpn}, trained and evaluated on ADE20K \cite{ade20k}. Backbone latency is measured under resolution $512\times512$. Models are grouped by mIoU. Best value in each group is made bold.  }
\label{tab:segmentation}
  
\end{minipage}%
\vspace{-0.2cm}
\end{figure*}

\paragraph{Conclusion.}
Reducing the kernel size of a convolution from $2\times2$ to $3\times3$, leads to a reduction of MACs by approximately 50\%. On CPU, this advantage is retained, as the MACpS remain similar (depthwise) or even improve (FuMBConv \& GrFuMBConv). However on GPU, similarly to grouping, reducing the kernel size leads to minimal latency benefits or even an increase in latency, as the MACpS deteriorates.

\vspace{-0.2cm}
\section{CPUBone}
\vspace{-0.1cm}
\label{sec:cpubone}
The macro architecture of CPUBone is inspired by LowFormer \cite{nottebaum2025lowformer,nottebaum2026macshardwareefficientarchitecture}, as it represents one of the most recent vision backbone approaches incorporating MBConv blocks as a main component. Following \cite{nottebaum2025lowformer,nottebaum2026macshardwareefficientarchitecture}, we also integrate LowFormer Attention in the final two stages of CPUBone.
However, based on the findings in Section \ref{sec:mbconvblock}, we exclusively use the grouped MBConv variants (GrMBConv and GrFuMBConv), applying the fused version when the input channel dimension is below $256$ and the unfused version otherwise.
We further only reduce the kernel size to $2\times2$ in the last two stages of CPUBone, where the bulk of computation is concentrated. We believe higher kernel size is especially important in earlier layers to have a higher receptive field and consequently better accuracy. In contrast to the LowFormer architecture, we also omit the transpose convolution in the LowFormer Attention and replace it with nearest neighbour upsampling. We ablate this decision in Subsection \ref{subsec:ablation}. 
In total, we feature four model variants with increasing model size: CPUBone-B0, B1, B2 and B3.
The overall architecture is illustrated in Figure \ref{fig:architecture} and specifics to each model variant are listed in Table \ref{tab:arch_specs}.

\section{Experiments}
\label{sec:experiments}
\subsection{ImageNet Classification}
\vspace{-0.1cm}
\paragraph{Settings.}
We perform image classification experiments on ImageNet-1K \cite{imagenet}, which consists of 1.28M training and 50K validation across for 1000 categories. All CPUBone models were trained from scratch using mostly the same setting as \cite{nottebaum2025lowformer, efficientvit} and featuring an input resolution of 224 for evaluation. We train for a total of 320 epochs using AdamW \cite{adamwpaper} optimizer, including 20 warm-up epochs with a linear schedule. We employ cosine decay \cite{cosinelrpaper} as learning rate scheduler. For CPUBone-B0 and B1, we use a base learning rate of $10^{-3}$ and a batch size of 512. For CPUBone-B2, the batch size is increased to 1024. For CPUBone-B3, we use a batch size of 2400 and raise the base learning rate to $3\times 10^{-3}$.

\paragraph{Results.}

In Table \ref{tab:imagenetresults} we compare the CPUBone models to recent vision backbones.
We compare model efficiency by measuring latency on the ARM CPU of the Raspberry Pi5, the CPU of the Pixel Pro 7 and the Intel(R) Xeon(R) W-2125 CPU, with a batch size of 1.
Regarding results, CPUBone-B0 is faster than almost all models with a lower accuracy in Table \ref{tab:imagenetresults}. 
EffFormerV2-S0 \cite{efficientformerv2} for example has a 50\% higher Pi5 latency, a 300\% higher Pixel 7 Pro CPU latency, but a 3.9\% lower accuracy, than CPUBone-B0. On the other side, FasterNet-T1 \cite{fasternet}, an architecture partially designed for CPU latency, achieves a slightly lower Intel CPU latency than CPUBone-B0, however its Pi5 CPU latency is 33\% higher,  its Pixel 7 Pro CPU latency is 76\% higher and its top-1 accuracy is 1.4\% lower.
On the other end of the spectrum regarding model size CPUBone-B3 is considerably faster than RepViT-M2.3 \cite{repvit} on all three CPU devices, while being 0.6\% more accurate on ImageNet \cite{imagenet}. 
Overall, CPUBone models consistently achieve efficient performance across various model sizes and a diverse set of CPU platforms.
 

\begin{table*}[t]
\centering
\small
    \begin{tabular}{c|cc|c|c|c|c|c}
        \Xhline{1.5pt}
         \multirow{2}{*}{Model version}  & Params &  MACs & \multicolumn{1}{c|}{Pi5 CPU Lat.} & Pixel 4 CPU &  \multicolumn{1}{c|}{Pixel 7 Pro CPU Lat.}& \multicolumn{1}{|c|}{Intel CPU Lat.}  & Top-1   \\
         & (M) & (M) & (ms) &  (ms) & (ms) &  (ms) &  (\%)  \\

       



        \Xhline{1.0pt}
        
        \textbf{Just original MBConvs} & 13.3 & 758 &  48.6   & 25.4  &  20.1  & 12.63  & 78.5 \footnotesize \textbf{(-0.2\%)} \\
        
        \textbf{Using transpose} & 14.4 &  840 &  39.7  & 26.3 &  20.3  & 10.73  & 78.9 \footnotesize \textbf{(+0.2\%)} \\
        
        \textbf{{Groups=8}}  & 11.0 & 560 &  28.7   & 21.2 & 16.0  & 8.8  &  78.0 \footnotesize \textbf{(-0.7\%)}\\
        \textbf{{Groups=4}}  & 11.5 & 622 &  29.6   & 21.9 & 17.9  & 9.0  &  78.4 \footnotesize \textbf{(-0.3\%)}\\

        \textbf{B0-plain} & 14.1 & 944 &  39.1  & 29.6  &   20.8  & 9.4 & 78.4 \footnotesize \textbf{(-0.3\%)} \\
        \Xhline{1.0pt}
        \textbf{Baseline (B1)} & 12.4 & 746 &  33.5  & 23.8 &  18.6  & 9.4  & 78.7  \\
        
        \Xhline{1.5pt}
     
    \end{tabular}

    \caption{Ablation study of CPUBone-B1, featuring singular changes. We also include B0-plain, an ablation to CPUBone-B0.  }
    \label{tab:ablation}
\vspace{-0.1cm}
\end{table*}

\subsection{Downstream Tasks}

To assess transferability of CPUBone backbones to downstream tasks, we integrate the pretrained models into object detection and semantic segmentation frameworks. Experiments are performed on COCO 2017 \cite{cocopaper} and ADE20K \cite{ade20k} using the MMDetection \cite{mmdetection} and MMSegmentation \cite{mmseg2020} toolkits. 
We use RetinaNet \cite{retinanet} for detection and Semantic FPN \cite{semanticfpn} for segmentation. Results are depicted in Tables   \ref{tab:segmentation} and \ref{tab:detection}. Pi5 CPU latency refers to backbone latency measured under resolution 512$\times$512.


\begin{table}[h]
    \centering
    \small
\begin{tabular}{c|c|c}
        \Xhline{1.5pt}
         \multirow{2}{*}{Backbone}  & \multicolumn{1}{c|}{Pi5 CPU Lat.}   &AP  \\
         & (ms) & (\%)  \\
         \Xhline{1.0pt}

        MobileNetV3 \cite{mobilenetv3}  & 158.0  & 29.9  \\

        MNv4-Conv-M \cite{mobilenetv4} & 299.5  & 32.6 \\
        PVTv2-B0  \cite{pvtv2} & 587.7   & 37.2  \\

       \rowcolor{lightergray}   CPUBone-B0 (ours)  & \textbf{131.5}   & \textbf{37.5}\\
        \Xhline{1.0pt}
        LowFormer-B0 \cite{nottebaum2025lowformer} & 226.9  & 38.6 \\
        EdgeViT-XXS \cite{edgevit}  & 281.0  & 38.7   \\

        \rowcolor{lightergray} CPUBone-B1 (ours)  & \textbf{189.9}  & \textbf{39.0}  \\
        \Xhline{1.0pt}
     LowFormer-B1 \cite{nottebaum2025lowformer} & \textbf{313.2}  & 39.4 \\
        
        FAT-B0 \cite{fat}  & 763.3   & 40.4  \\


      \rowcolor{lightergray} CPUBone-B2 (ours) & 338.2  & \textbf{40.4} \\
        \Xhline{1.0pt}
        EdgeViT-XS \cite{edgevit}  & 461.5   & 40.6   \\
        
        PVTv2-B1  \cite{pvtv2} & 1296.4  & 41.2  \\
        LowFormer-B2 \cite{nottebaum2025lowformer} & \textbf{808.1}    & 41.4 \\
        FAT-B1 \cite{fat}  & 1102.1  & 42.5   \\
       
        \rowcolor{lightergray} CPUBone-B3 (ours) & 1181.6   & \textbf{42.9}  \\

     
     \Xhline{1.5pt}
    \end{tabular}
    \caption{Results on object detection using RetinaNet head \cite{retinanet}. Backbone latency is measured under resolution $512\times512$.  Models are grouped by AP. Best value in each group is made bold.}
\label{tab:detection}
\vspace{-0.2cm}
\end{table}

\paragraph{Semantic Segmentation.}
For semantic segmentation, we train the models for 40K iterations with a batch size of 32, following the protocols in \cite{fat, repvit, fastvit, efficientmodulation}. 
Regarding results, CPUBone consistently achieves lower latency and higher mIoU across a wide range of model sizes (see Table \ref{tab:segmentation}). CPUBone executes up to $3\times$ faster than comparable models with similar or lower mIoU.

\paragraph{Object Detection.}
We train all models for 12 epochs using the standard 1× schedule, following the setup in \cite{efficientvit, fat}.  Similarly to semantic segmentation, CPUBone is able to outperform all compared models with a similar AP (see Table \ref{tab:detection}) and executes up to $4\times$ faster than comparable models with a similar or lower AP.


\subsection{Ablation Study}
\label{subsec:ablation}
To verify that our design choices yield a CPU-optimal model, we perform the following ablations on CPUBone-B1 in Table \ref{tab:ablation}: 
 setting \textbf{groups=4} instead of two in all MBConv blocks, setting \textbf{groups=8} instead of two in all MBConv blocks, featuring \textbf{just original MBConvs} \cite{mobilenetv2} instead of grouped or fused variants and \textbf{using transpose} convolutions in LowFormer Attention \cite{nottebaum2025lowformer}, as originally designed, instead of nearest neighbour upsampling.
  Finally, we also ablate CPUBone-B0 in \textbf{B0-plain}, where we revert each of our contributions: grouping, reduced kernel size, and the replacement of the transpose convolution with nearest neighbour upsampling.
\vspace{0.15cm}
\newline
\textbf{B0-plain} does have a increased accuracy, compared to CPUBone-B0, however when we directly compare it to CPUBone-B1, it not only is considerably  slower on all devices, but also has a lower top-1 accuracy.
\vspace{0.15cm}
\newline
\textbf{Groups=4 or Groups=8} on the other side, improve latency due to a lower MAC count, however lead to a considerable reduction in accuracy. Higher group numbers  further lead to an increasingly declining hardware-efficiency.
On the ARM CPU, the \textbf{Groups=4} model shows a 5\% reduction in MACpS, while \textbf{Groups=8} shows a 13\% reduction relative to the baseline. 
Further, we can see that the efficiency benefit on Intel CPU is minimal for those models. Our baseline (B1) with $groups$ set to two is therefore a good balance between MAC count and hardware-efficiency (MACpS).
\vspace{0.15cm}
\newline
\textbf{Using transpose} convolutions in LowFormer Attention improves accuracy by 0.2\%, but it increases latency considerably on all CPU devices. On ARM CPU it further leads to 5\% lower MACpS, compared to the Baseline.
\vspace{0.15cm}
\newline
\textbf{Just original MBConv} blocks instead of GrMBConv or GrFuMBConv blocks 
 lead to a significantly lower accuracy and degrades latency on all devices. This is the case, even though it has a similar MAC count as our Baseline B1, demonstrating the enormous hardware-efficiency benefit of our design by optimizing MACpS.

\section{Conclusion}

In this work, we investigated strategies for CPU-efficient model design.
%
We pointed out, that a major limitation for CPUs is their low parallelization capability, making them less suited for models with a high MAC count.
A common approach to efficient model design is the use of depthwise convolutions. While they  feature a comparably low MAC count, they also suffer from low MACpS, even on CPU devices.
%
To address this, we proposed two alternative design adaptations for standard convolutions: setting the number of groups to two and reducing the kernel size.
%
%
We explicitly computed the MAC reduction for both modifications and verified experimentally that the hardware-efficiency of the convolution, measured in MACpS, is maintained during execution on CPU.
%
%
%
%
%
Building on these insights, we introduced CPUBone, a new vision backbone architecture family, specifically optimized for CPU-based inference.
At its core, the Grouped Fused MBConv (GrFuMBConv) block effectively combines low MAC count and high MACpS.
CPUBone consistently outperforms existing models across a wide range of CPU devices and effectively transfers its efficiency to downstream tasks such as object detection and semantic segmentation.

\paragraph{Acknowledgements.}This research has been funded by the European Union, NextGenerationEU – PNRR M4 C2 I1.1, RS Micheloni. Progetto PRIN 2022 PNRR - “Tracking in Egovision for Applied Memory (TEAM)” Code P20225MSER 001 Code CUP G53D23006680001. Matteo Dunnhofer received funding from the European Union’s
Horizon Europe research and innovation programme un-
der the Marie Skłodowska-Curie grant agreement n.
101151834 (PRINNEVOT CUP G23C24000910006).
We also acknowledge ISCRA for awarding this project access to the LEONARDO supercomputer, owned by the EuroHPC Joint Undertaking, hosted by CINECA (Italy).

{\small
\bibliographystyle{ieeenat_fullname}
\bibliography{main}
}

\clearpage
\setcounter{page}{1}
\maketitlesupplementary


\section{Additional Details to CPU-Efficient Design Strategies}
\label{supp:sec:hardwarefff}

\subsection{MAC calculations}

With Equation~\eqref{eq:maccount_general} we can calculate the MACs of a convolution with no bias. This further allows us to compute the reduction in MACs of the grouped MBConv variants (GrMBConv \& GrFuMBConv) compared to their ungrouped counterparts (MBConv \& FuMBConv), which we needed in Subsection \ref{subsubsec:grouping_hardwareeff}. As well as calculate how the reduction of the kernel size from   $3\times3$ to $2\times2$ affects MACs of the whole FuMBConv and GrFuMBConv block, which we need in Subsection \ref{subsubsec:kernel_hardwareeff}.
For MAC calculations, we leave out any batch normalization and activation functions of the MBConv blocks \cite{mobilenetv2}, due to having a minor influence on the MAC count. We further assume for simplicity, that the input channel dimension ($C_{in}$) and the output channel dimension ($C_{out}$) are the same, that the expansion factor is 4 and the stride is 1.

\subsubsection{FuMBConv vs. GrFuMBConv}

Following  Equation~\eqref{eq:maccount_general}, the MACs of the fused MBConv block ($M_{fmb}$) or the GrFuMBConv block can be computed as

\begin{equation}\label{supp:eq:maccount_fumbconv}
\begin{aligned}
   M_{fmb} =& \frac{K_{H} \times K_W \times C_{in} \times C_{in} \times  H_{out} \times W_{out} }{groups}   \\
   +& C_{in} \times C_{in} \times 4 \times H_{out} \times W_{out},
\end{aligned}
\end{equation}
where the first part refers to the standard convolution and the second part to the pointwise convolution (see Figure \ref{fig:mbconvs_design}).
By pulling out common multiplicative factors, Equation~\eqref{supp:eq:maccount_fumbconv} can be simplified as:
\begin{equation}\label{supp:eq:maccount_fumbconv_simple}
\begin{aligned}
   M_{fmb} =&  (C_{in} \times C_{in} \times 4 \times H_{out} \times W_{out}) \times \\
   &(K_{H} \times \frac{K_W}{groups} + 1),
\end{aligned}
\end{equation}
where the term of the pointwise convolution is reduced to a single number in the second bracket.
The only distinction between FuMBConv \cite{fusedmbconv} and GrFuMBConv in Equation ~\eqref{supp:eq:maccount_fumbconv} lies in the $groups$ parameter. Therefore, Equation ~\eqref{supp:eq:maccount_fumbconv_simple} allows us to compute the proportion of MACs that GrFuMBConv requires relative to FuMBConv.
When having a kernel of size $3\times3$ (and $groups=1$ for the FuMBConv and $groups=2$ for the GrFuMBConv, while all other parameters are the same), the MAC count of GrFuMBConv divided by FuMBConv is:
\begin{equation}\label{supp:eq:rel_maccount_fumbconvgr_1}
\begin{aligned}
     =&  \frac{(C_{in} \times C_{in} \times 4 \times H_{out} \times W_{out})}{(C_{in} \times C_{in} \times 4 \times H_{out} \times W_{out})} \times \\
   &\frac{(3 \times \frac{3}{2} + 1)}{(3 \times \frac{3}{1} + 1)}, 
\end{aligned}
\end{equation}
which can be further simplified to:
\begin{equation}\label{supp:eq:rel_maccount_fumbconvgr_2}
\begin{aligned}
     =&  \frac{1}{1} \times  \frac{5.5}{10} = 0.55 \quad.
\end{aligned}
\end{equation}
Consequently, the GrFuMBConv always has 45\% less MACs than its ungrouped counterpart (when they only differ in the $groups$ paramter), independent of the channel dimension.

\subsubsection{ MBConv vs GrMBConv}
The original MBConv block \cite{mobilenetv2} consists out of two pointwise convolutions and one depthwise convolution (see Figure \ref{fig:mbconvs_design}). Following Equation~\eqref{eq:maccount_general}, we can calculate the MAC count of the MBConv ($M_{mb}$) and GrMBConv block (only differing in the $groups$ parameter) as follows:

\begin{equation}\label{supp:eq:maccount_origmbconv}
\begin{aligned}
   M_{mb} =& \frac{  C_{in} \times C_{in} \times 4 \times  H_{out} \times W_{out} }{groups}   \\
   +& C_{in} \times 4 \times K_H \times K_W \times H_{out} \times W_{out}  \\
   +& C_{in} \times C_{in} \times 4 \times H_{out} \times W_{out}.
\end{aligned}
\end{equation}
Similarly to what we did in Equation~\eqref{supp:eq:maccount_fumbconv_simple}, we can simplify Equation~\eqref{supp:eq:maccount_origmbconv} by pulling out the commmon multiplicative factors, leaving us with this:

\begin{equation}\label{supp:eq:maccount_origmbconv_simple1}
\begin{aligned}
   M_{mb} =&  C_{in} \times 4 \times H_{out} \times W_{out}  \times \\
   &(\frac{  C_{in}  }{groups}      +   K_H \times K_W   +  C_{in} ).
\end{aligned}
\end{equation}
Similarly to Equation~\eqref{supp:eq:rel_maccount_fumbconvgr_1}, we can now easily compute the relative MAC count of GrMBConv, compared to MBConv, by dividing the number of MACs of GrMBConv ($groups=2$) with the MAC count of MBConv (with kernel size set to $3\times3$ for both):

\begin{equation}\label{supp:eq:maccount_origmbconv_relative}
\begin{aligned}
    =&  \frac{C_{in} \times 4 \times H_{out} \times W_{out}}{C_{in} \times 4 \times H_{out} \times W_{out}}  \times \\
   & \frac{(\frac{  C_{in}  }{2}      +   3 \times 3   +  C_{in} )}{(\frac{  C_{in}  }{1}      +   3 \times 3   +  C_{in} )}, 
\end{aligned}
\end{equation}
which can be further simplified to:
\begin{equation}\label{supp:eq:maccount_origmbconv_relative_simple1}
\begin{aligned}
   =&  \frac{1}{1}  \times  \frac{(\frac{  C_{in}  }{2}      +   9   +  C_{in} )}{(\frac{  C_{in}  }{1}      +   9   +  C_{in} )} 
   = \frac{(   1.5 \times C_{in} +  9    )}{( 2 \times C_{in}      +   9    )} \\
   &= \frac{1.5}{2} \times \frac{C_{in} + 6}{ C_{in} + 4.5}.
\end{aligned}
\end{equation}
Equation~\eqref{supp:eq:maccount_origmbconv_relative_simple1} above shows us, that the relative MAC count of GrMBConv divided by MBConv, depends on the channel dimension. However it further shows that with increasing $C_{in}$, it converges to $1.5/2=0.75$, meaning the grouped MBConv variant has 25\% less MACs than its ungrouped counterpart.
For the channel dimensions in Table \ref{tab:mbconv_eff_exp}, namely 32, 64, 128, 256 and 512, the relative MACs in percentage rounded accordingly are, respectively: 78.1\%, 76.6\%, 75.8\%, 75.4\%, 75.2\%. The average is 76.2\%.

\subsubsection{Impact of Kernel Size Reduction on MAC Count for GrFuMBConv and FuMBConv}

In Subsection \ref{subsubsec:kernel_hardwareeff} we compared how a reduction of the kernel size from $3\times3$ to $2\times2$ affects MACpS for the depthwise convolution, the FuMBConv and the GrFuMBConv with $groups=2$. In Subsection \ref{subsec:mac_calculation} we calculated for convolutions in general, how the reduced kernel affects MAC count, leading to approximately 56\% always. Therefore the MAC count of the depthwise convolution is reduced by 56\%, when reducing the kernel size.
However, the FuMBConv block and GrFuMBConv block also include an additional pointwise convolution, whose kernel is not reduced, making the calculation a bit more complicated. However, we need the MAC count of the full MBConv blocks, since we also tested the full MBConv block in Subsection \ref{subsubsec:kernel_hardwareeff}.

Starting from Equation~\eqref{supp:eq:maccount_fumbconv_simple}, which describes the MAC count of the GrFuMBConv or the FuMBConv, we can divide the MACs of the $2\times2$ FuMBConv  by the $3\times3$ FuMBConv block to obtain the percentage reduction (similarly to what we did in Equation~\eqref{supp:eq:rel_maccount_fumbconvgr_1}): 

\begin{equation}\label{supp:eq:fugrmbconv_kernelreduc_calcfumbconv}
\begin{aligned}
   =&  \frac{(C_{in} \times C_{in} \times 4 \times H_{out} \times W_{out})}{(C_{in} \times C_{in} \times 4 \times H_{out} \times W_{out})} \times \\
   & \frac{2 \times \frac{2}{1} + 1}{3 \times \frac{3}{1} + 1} \\
   =&  \frac{1}{1} \times  \frac{5}{10} = 0.5  .
\end{aligned}
\end{equation}
Consequently, the kernel reduction does exactly halve the MAC count of the FuMBConv block.
By doing the same thing now for the GrFuMBConv block, we yield:

\begin{equation}\label{supp:eq:fugrmbconv_kernelreduc_calcgrfumbconv}
\begin{aligned}
   =&  \frac{(C_{in} \times C_{in} \times 4 \times H_{out} \times W_{out})}{(C_{in} \times C_{in} \times 4 \times H_{out} \times W_{out})} \times \\
   & \frac{2 \times \frac{2}{2} + 1}{3 \times \frac{3}{2} + 1} \\
   =&  \frac{1}{1} \times  \frac{3}{5.5} = 0.\overline{54} \quad,
\end{aligned}
\end{equation}
 meaning for the GrFuMBConv block the kernel reduction leads to approximately 46\% less MACs.


\subsection{Kernel Experiment GPU }
In Table \ref{supp:tab:kernel_exp_depthwise_gpu} \& \ref{supp:tab:kernel_exp_groups2_fused_gpu}, we repeat the experiments of Subsection \ref{subsubsec:kernel_hardwareeff} from table \ref{tab:kernel_exp_depthwise} \& \ref{tab:kernel_exp_groups2_fused}, but on GPU instead of the ARM CPU of the Raspberry Pi5.
Table \ref{supp:tab:kernel_exp_depthwise_gpu} \& \ref{supp:tab:kernel_exp_groups2_fused_gpu} show an extreme deterioration of the MACpS, when reducing the kernel of a convolution from $3\times3$ to $2\times2$. Especially for depthwise convolutions, this change leads to a higher execution time than compared to the original $3\times3$ kernel. For FuMBConv and GrFuMBConv, the MACpS also decrease significantly; however, both variants retain at least a similar latency compared to the original $3\times3$ kernel. While they achieve roughly half the MACs, the corresponding reduction in MACpS offsets the expected efficiency gain, effectively nullifying it.

\begin{table}[H]
\centering
\setlength{\tabcolsep}{6.5pt}
\renewcommand{\arraystretch}{1.25}
\footnotesize
\begin{tabular}{cc|cccc|c}
    \Xhline{1.5pt}
    \multicolumn{2}{c|}{\textbf{Dwise GPU}} & \multicolumn{4}{c|}{\textbf{Channel Dimension}} & \\[-3pt]
    \textbf{Resol.} & \textbf{Type} & 128 & 256 & 512 & 1024 & \textbf{Avg.} \\
    \Xhline{1.5pt}

    \multirow{2}{*}{\textbf{7×7}} 
        & \textit{\textcolor{gray}{nmk}} & \textbf{6.6} & \textbf{13.1} & \textbf{26.5} & \textbf{52.4} & \textbf{24.7} \\[-2pt]
        & \textit{\textcolor{gray}{smk}} & 1.2 & 2.3 & 4.7 & 9.3 & 4.4 \\
    \cline{1-7}

    \multirow{2}{*}{\textbf{14×14}} 
        & \textit{\textcolor{gray}{nmk}} & \textbf{26.4} & \textbf{52.7} & \textbf{103.8} & \textbf{205.3} & \textbf{97.0} \\[-2pt]
        & \textit{\textcolor{gray}{smk}} & 4.7 & 9.0 & 18.7 & 36.3 & 17.2 \\

      \cline{1-7}

    \multirow{2}{*}{\textbf{28×28}} 
        & \textit{\textcolor{gray}{nmk}} & \textbf{103.2} & \textbf{203.5} & \textbf{300.3} & \textbf{308.0} & \textbf{228.7} \\[-2pt]
        & \textit{\textcolor{gray}{smk}} & 18.1 & 35.7 & 71.5 & 68.9 & 48.5 \\

    \Xhline{1.5pt}
\end{tabular}

\caption{Execution efficiency experiment on GPU on the effect of kernel size reduction for depthwise convolutions, by measuring MMACs/ms. \textbf{Resol.} refers to operating resolution, \textbf{Type} refers to the kernel size ($nmk=3\times3$, $smk=2\times2$). Bold entries indicate whether nmk or smk variant has higher MACpS. Same as Table \ref{tab:kernel_exp_depthwise}, but on GPU.}
\label{supp:tab:kernel_exp_depthwise_gpu}
\end{table}

\begin{table}[H]
\centering
\setlength{\tabcolsep}{5pt}
\renewcommand{\arraystretch}{1.25}
\footnotesize
\resizebox{8cm}{!}{
\begin{tabular}{cl|cc|c!{\vrule width 1.5pt}cc|c}
\Xhline{1.5pt}
 \multicolumn{2}{c|}{\textbf{GPU}} & \multicolumn{3}{c!{\vrule width 1.5pt}}{\textbf{Ungrouped}} & \multicolumn{3}{c}{\textbf{Groups=2}} \\
\cline{3-8}
\textbf{Resol.} & \textbf{Type} & \multicolumn{2}{c|}{\textbf{Channel Dim.}} &  & \multicolumn{2}{c|}{\textbf{Channel Dim.}} & \\
\cline{3-4} \cline{6-7}
 & & 128 & 256 &\textbf{Avg.} & 128 & 256 & \textbf{Avg.} \\
\Xhline{1.5pt}

\multirow{2}{*}{\textbf{7×7}}
 & \textit{\textcolor{gray}{nmk}} & \textbf{661.6} & \textbf{1075.7} & \textbf{868.6 }& \textbf{355.1} & \textbf{893.7} & \textbf{624.4} \\[-2pt]
 & \textit{\textcolor{gray}{smk}} & 284.4 & 882.9 & 583.6 & 145.3 & 627.2 & 386.2 \\
\cline{1-8}

\multirow{2}{*}{\textbf{14×14}}
 & \textit{\textcolor{gray}{nmk}} & \textbf{2374.3} & \textbf{3545.5} & \textbf{2959.9} & \textbf{1422.6} & \textbf{3026.8} & \textbf{2224.7} \\[-2pt]
 & \textit{\textcolor{gray}{smk}} & 1081.8 & 2021.3 & 1551.5 & 627.0 & 1671.9 & 1149.4 \\
\cline{1-8}

\multirow{2}{*}{\textbf{28×28}}
 & \textit{\textcolor{gray}{nmk}} & \textbf{4892.2} & \textbf{5385.9} & \textbf{5139.0} & \textbf{4068.5} & \textbf{4785.6} & \textbf{4427.0} \\[-2pt]
 & \textit{\textcolor{gray}{smk}} & 2333.9 & 2669.0 & 2501.5 & 2000.5 & 2538.2 & 2269.3 \\
\Xhline{1.5pt}
\end{tabular}}

\caption{Execution efficiency experiment on GPU on the effect of kernel size reduction. We test the GrFuMBConv (Groups=2) and FuMBConv (Ungrouped) block by measuring MMACs/ms.  \textbf{Resol.} refers to operating resolution, \textbf{Type} refers to the kernel size ($nmk=3\times3$, $smk=2\times2$). Bold entries indicate whether nmk or smk variant has higher MACpS. Same as Table \ref{tab:kernel_exp_groups2_fused}, but on GPU. }
\label{supp:tab:kernel_exp_groups2_fused_gpu}
\end{table}

\begin{table*}[ht]
\centering
\small
\setlength{\tabcolsep}{6pt}
\renewcommand{\arraystretch}{1.25}

\begin{tabular}{c|l|ccccc|c}
\Xhline{1.5pt}
\multicolumn{2}{c|}{\textbf{Grouping=2 Experiment ARM CPU}} & \multicolumn{5}{c|}{\textbf{Channel Dimension}} & \\
\cline{3-7}

\multicolumn{1}{c}{\textbf{Resolution}} & \multicolumn{1}{c|}{\textbf{Variant}} & 32 & 64 & 128 & 256 & 512 & \textbf{Avg.} \\
\Xhline{1.5pt}

    & \textit{\textcolor{gray}{MBConv}} & \textbf{4.1} & \textbf{8.5} & \textbf{14.9} & \textbf{22.4} & \textbf{23.5} & 14.7 \\[-2pt]
  \textbf{7x7}  & \textit{\textbf{Gr}\textcolor{gray}{MBConv}} & 3.7 & 6.5 & 11.6 & 17.7 & 19.5 & 11.8 \footnotesize \textbf{(-19\%)} \\ 
    \cline{2-8}
 (MMACs/ms)   & \textit{\textbf{Fu}\textcolor{gray}{MBConv}} & \textbf{25.5} & \textbf{36.3} & \textbf{38.7} & 27.8 & 24.3 & 30.5 \\[-2pt]
    & \textit{\textbf{GrFu}\textcolor{gray}{MBConv}} & 14.7 & 28.3 & 37.2 & \textbf{32.3} & \textbf{26.5} & 27.8 \footnotesize \textbf{(-8\%)} \\
\Xhline{1.2pt}

    & \textit{\textcolor{gray}{MBConv}} & \textbf{7.8} & \textbf{13.3} & \textbf{21.4} & \textbf{26.4} & \textbf{26.2} & 19.0 \\[-2pt]
  \textbf{14x14}  & \textit{\textbf{Gr}\textcolor{gray}{MBConv}} & 6.3 & 10.6 & 16.7 & 22.1 & 23.3 & 15.8 \footnotesize \textbf{(-16\%)} \\ 
    \cline{2-8}
 (MMACs/ms)   & \textit{\textbf{Fu}\textcolor{gray}{MBConv}} & \textbf{40.6} & \textbf{45.3} & 40.7 & 32.4 & 24.5 & 36.7 \\[-2pt]
    & \textit{\textbf{GrFu}\textcolor{gray}{MBConv}} & 31.0 & 42.1 & \textbf{44.4} & \textbf{34.7} & \textbf{30.4} & 36.5 \footnotesize \textbf{(-0\%)} \\
\Xhline{1.2pt}

   & \textit{\textcolor{gray}{MBConv}} & \textbf{11.7} & \textbf{18.6} & \textbf{22.5} & \textbf{26.3} & \textbf{26.9} & 21.2 \\[-2pt]
  \textbf{28x28}  & \textit{\textbf{Gr}\textcolor{gray}{MBConv}} & 9.4 & 14.8 & 18.1 & 21.3 & 26.8 & 18.1  \footnotesize \textbf{(-14\%)}\\ 
    \cline{2-8}
 (MMACs/ms)   & \textit{\textbf{Fu}\textcolor{gray}{MBConv}} & \textbf{53.2} & \textbf{54.6} & 41.2 & 30.0 & 25.9 & 41.0 \\[-2pt]
    & \textit{\textbf{GrFu}\textcolor{gray}{MBConv}} & 45.5 & 53.4 & \textbf{44.1} & \textbf{37.5} & \textbf{29.2} & 41.9 \footnotesize \textbf{(+2\%)} \\
\Xhline{1.2pt}

   & \textit{\textcolor{gray}{MBConv}} & \textbf{8.4} & \textbf{12.8} & \textbf{19.1} & \textbf{22.7} & \textbf{30.8} & 18.8 \\[-2pt]
  \textbf{56x56}  & \textit{\textbf{Gr}\textcolor{gray}{MBConv}} & 7.2 & 9.9 & 14.3 & 18.4 & 25.4 & 15.0 \footnotesize \textbf{(-20\%)}\\ 
    \cline{2-8}
 (MMACs/ms)   & \textit{\textbf{Fu}\textcolor{gray}{MBConv}} & \textbf{42.3} & \textbf{37.5} & 31.7 & 24.4 & 23.2 & 31.8 \\[-2pt]
    & \textit{\textbf{GrFu}\textcolor{gray}{MBConv}} & 35.3 & 37.3 & \textbf{34.8} & \textbf{27.2} & \textbf{24.0} & 31.7 \footnotesize \textbf{(-0\%)} \\
\Xhline{1.2pt}

\Xhline{1.5pt}
\end{tabular}

\caption{Execution efficiency of MACs, measured in MMACs/ms on the ARM CPU of the Raspberry Pi5, for the four MBConv variants (MBConv, GrMBConv, FuMBConv and GrFuMBConv) for different channel dimensions and operating resolutions. Bold entries indicate whether the grouped or ungrouped variant has higher MACpS. All grouped variants have $groups=2$. It is similar to Table \ref{tab:mbconv_eff_exp}, but for resolutions $7\times7$, $14\times14$, $28\times28$ and $56\times56$ and only on ARM CPU.}
\label{supp:tab:mbconv_eff_exp_armcpu}
\end{table*}

\subsection{Grouping ARM CPU - additional Resolutions}

Table \ref{supp:tab:mbconv_eff_exp_armcpu} shows a similar experiment as Table \ref{tab:mbconv_eff_exp}, however we focus only on ARM CPU, but feature the additional resolutions $7\times7$, $28\times28$ and $56\times56$.
The numbers are mostly very similar to \ref{tab:mbconv_eff_exp}, independent of the resolutions. However the GrMBConv consistently underperforms on ARM CPU, compared to the other CPU devices featured in Table \ref{tab:mbconv_eff_exp} (Snapdragon 8 Elite CPU, Google Pixel 4 CPU). The FuMBConv block however consistently shows high MACpS, only on resolution $7\times7$ it falls off a bit. 

\subsection{Grouping 4 on ARM CPU}

In Table \ref{supp:tab:mbconv_eff_exp_armcpu_group4} we repeat the experiment of Table \ref{tab:mbconv_eff_exp} for additional resolutions ($7\times7$, $14\times14$, $28\times28$ and $56\times56$) and on the ARM CPU, but for $groups=4$ instead of $groups=2$. The averaged numbers with $groups=4$ (Table \ref{supp:tab:mbconv_eff_exp_armcpu_group4}) are very similar to $groups=2$ (Table \ref{tab:mbconv_eff_exp}), however the distribution over the channels is  different: For channel dimensions below 128,  GrFuMBConv with $groups=4$ has 20\% lower MACpS, compared to $groups=2$. However for channel dimensions over 128, they have 36\% higher MACpS. Consequently, depending on the channel dimension, either $groups=2$ (below 128) or $groups=4$ (over 128) is more hardware efficient. 

\begin{table}[ht]
\centering
\small
\setlength{\tabcolsep}{6pt}
\renewcommand{\arraystretch}{1.25}
\resizebox{8cm}{!}{
\begin{tabular}{c|l|ccccc|c}
\Xhline{1.5pt}
\multicolumn{2}{c|}{\textbf{Grouping=4 Experiment ARM CPU}} & \multicolumn{5}{c|}{\textbf{Channel Dimension}} & \\
\cline{3-7}

\multicolumn{1}{c}{\textbf{Resolution}} & \multicolumn{1}{c|}{\textbf{Variant}} & 32 & 64 & 128 & 256 & 512 & \textbf{Avg.} \\
\Xhline{1.5pt}

   & \textit{\textcolor{gray}{MBConv}} & \textbf{4.1} & \textbf{8.5} & \textbf{14.9} & \textbf{22.4} & \textbf{23.5} & 14.7 \\[-2pt]
  \textbf{7x7}  & \textit{\textbf{Gr}\textcolor{gray}{MBConv}} & 3.1 & 5.4 & 10.2 & 17.4 & 19.9 & 11.2 \footnotesize \textbf{(-23\%)} \\
\noalign{\setlength{\arrayrulewidth}{0.1pt}}
\cline{2-8}
\noalign{\setlength{\arrayrulewidth}{0.4pt}}
  (MMACs/ms)  & \textit{\textbf{Fu}\textcolor{gray}{MBConv}} & \textbf{25.5} & \textbf{36.3} & \textbf{38.7} & 27.8 & 24.3 & 30.5 \\[-2pt]
    & \textit{\textbf{GrFu}\textcolor{gray}{MBConv}} & 10.2 & 18.5 & 31.4 & \textbf{34.9} & \textbf{32.0} & 25.4 \footnotesize \textbf{(-16\%)}\\
\Xhline{1.2pt}

   & \textit{\textcolor{gray}{MBConv}} & \textbf{7.8} & \textbf{13.3} & \textbf{21.4} & \textbf{26.4} & 26.2 & 19.0 \\[-2pt]
  \textbf{14x14}  & \textit{\textbf{Gr}\textcolor{gray}{MBConv}} & 5.6 & 9.2 & 16.1 & 22.1 & \textbf{28.1} & 16.2  \footnotesize \textbf{(-14\%)}\\
\noalign{\setlength{\arrayrulewidth}{0.1pt}}
\cline{2-8}
\noalign{\setlength{\arrayrulewidth}{0.4pt}}
  (MMACs/ms)  & \textit{\textbf{Fu}\textcolor{gray}{MBConv}} & \textbf{40.6} & \textbf{45.3} & 40.7 & 32.4 & 24.5 & 36.7 \\[-2pt]
    & \textit{\textbf{GrFu}\textcolor{gray}{MBConv}} & 24.8 & 36.4 & \textbf{43.2} & \textbf{41.4} & \textbf{39.8} & 37.1 \footnotesize \textbf{(+1\%)} \\
\Xhline{1.2pt}

   & \textit{\textcolor{gray}{MBConv}} & \textbf{11.7} & \textbf{18.6} & \textbf{22.5} & \textbf{26.3} & 26.9 & 21.2 \\[-2pt]
  \textbf{28x28}  & \textit{\textbf{Gr}\textcolor{gray}{MBConv}} & 8.4 & 13.0 & 17.1 & 23.6 & \textbf{36.6} & 19.7 \footnotesize \textbf{(-7\%)} \\
\noalign{\setlength{\arrayrulewidth}{0.1pt}}
\cline{2-8}
\noalign{\setlength{\arrayrulewidth}{0.4pt}}
  (MMACs/ms)  & \textit{\textbf{Fu}\textcolor{gray}{MBConv}} & \textbf{53.2} & \textbf{54.6} & 41.2 & 30.0 & 25.9 & 41.0 \\[-2pt]
    & \textit{\textbf{GrFu}\textcolor{gray}{MBConv}} & 36.7 & 47.1 & \textbf{49.5} & \textbf{50.5} & \textbf{45.7} & 45.9 \footnotesize \textbf{(+12\%)} \\
\Xhline{1.2pt}

   & \textit{\textcolor{gray}{MBConv}} & \textbf{8.4} & \textbf{12.8} & \textbf{19.1} & \textbf{22.7} & 30.8 & 18.8 \\[-2pt]
  \textbf{56x56}  & \textit{\textbf{Gr}\textcolor{gray}{MBConv}} & 6.3 & 8.9 & 14.2 & 21.6 & \textbf{33.8} & 17.0 \footnotesize \textbf{(-9\%)}\\
\noalign{\setlength{\arrayrulewidth}{0.1pt}}
\cline{2-8}
\noalign{\setlength{\arrayrulewidth}{0.4pt}}
  (MMACs/ms)  & \textit{\textbf{Fu}\textcolor{gray}{MBConv}} & \textbf{42.3} & \textbf{37.5} & 31.7 & 24.4 & 23.2 & 31.8 \\[-2pt]
    & \textit{\textbf{GrFu}\textcolor{gray}{MBConv}} & 31.3 & 30.8 & \textbf{39.7} & \textbf{43.4} & \textbf{39.5} & 36.9 \footnotesize \textbf{(+16\%)} \\
\Xhline{1.2pt}

\Xhline{1.5pt}
\end{tabular}}

\caption{Execution efficiency of MACs, measured in MMACs/ms on the ARM CPU of the Raspberry Pi5, for the four MBConv variants (MBConv, GrMBConv, FuMBConv and GrFuMBConv) for different channel dimensions and operating resolutions. Bold entries indicate whether the grouped or ungrouped variant has higher MACpS. It is similar to Table \ref{tab:mbconv_eff_exp}, but for resolutions $7\times7$, $14\times14$, $28\times28$ and $56\times56$, only on ARM CPU and all grouped variants have $groups=4$ instead of $groups=2$ of Table \ref{tab:mbconv_eff_exp}.}
\label{supp:tab:mbconv_eff_exp_armcpu_group4}
\end{table}

\subsection{Grouping GPU - additional Resolutions}

In Table \ref{supp:tab:mbconv_eff_exp_gpu} we repeat the experiment of Table \ref{tab:mbconv_eff_exp} for additional resolutions ($7\times7$, $14\times14$, $28\times28$ and $56\times56$) on the Nvidia TITAN RTX GPU. We observe that at lower resolutions, the grouped variants perform worse, likely due to reduced opportunities for parallelizing the convolutional operations. From Table \ref{supp:tab:mbconv_eff_exp_gpu} we can conclude, that grouping convolutions does not fully translate lower MAC count to a similarly low latency on GPU.

\begin{table*}[ht]
\centering
\small
\setlength{\tabcolsep}{6pt}
\renewcommand{\arraystretch}{1.25}

\begin{tabular}{c|l|ccccc|c}
\Xhline{1.5pt}
\multicolumn{2}{c|}{\textbf{Grouping=2 Experiment GPU}} & \multicolumn{5}{c|}{\textbf{Channel Dimension}} & \\
\cline{3-7}

\multicolumn{1}{c}{\textbf{Resolution}} & \multicolumn{1}{c|}{\textbf{Variant}} & 32 & 64 & 128 & 256 & 512 & \textbf{Avg.} \\
\Xhline{1.5pt}

   & \textit{\textcolor{gray}{MBConv}} & \textbf{8.6} & \textbf{31.7} & \textbf{114.4} & \textbf{474.0} & \textbf{1746.1} & 475.0 \\[-2pt]
  \textbf{7x7}  & \textit{\textbf{Gr}\textcolor{gray}{MBConv}} & 5.6 & 20.4 & 82.4 & 328.4 & 1094.9 & 306.3 \footnotesize \textbf{(-35\%)} \\
\noalign{\setlength{\arrayrulewidth}{0.1pt}}
\cline{2-8}
\noalign{\setlength{\arrayrulewidth}{0.4pt}}
  (MMACs/ms)  & \textit{\textbf{Fu}\textcolor{gray}{MBConv}} & \textbf{45.6} & \textbf{182.4} & \textbf{672.7} & \textbf{1089.9} & \textbf{1309.8} & 660.1 \\[-2pt]
    & \textit{\textbf{GrFu}\textcolor{gray}{MBConv}} & 23.5 & 85.6 & 342.2 & 878.4 & 1264.3 & 518.8 \footnotesize \textbf{(-21\%)} \\
\Xhline{1.2pt}

   & \textit{\textcolor{gray}{MBConv}} & \textbf{34.2} & \textbf{112.3} & \textbf{428.6} & \textbf{1747.3} & \textbf{3494.0} & 1163.3 \\[-2pt]
  \textbf{14x14}   & \textit{\textbf{Gr}\textcolor{gray}{MBConv}} & 25.0 & 88.4 & 308.8 & 1260.6 & 2679.1 & 872.4 \footnotesize \textbf{(-25\%)} \\
\noalign{\setlength{\arrayrulewidth}{0.1pt}}
\cline{2-8}
\noalign{\setlength{\arrayrulewidth}{0.4pt}}
  (MMACs/ms)  & \textit{\textbf{Fu}\textcolor{gray}{MBConv}} & \textbf{185.1} & \textbf{622.6} & \textbf{2394.6} & \textbf{3612.2} & 3849.0 & 2132.7 \\[-2pt]
    & \textit{\textbf{GrFu}\textcolor{gray}{MBConv}} & 96.8 & 330.3 & 1307.3 & 2963.3 & \textbf{3878.9} & 1715.3 \footnotesize \textbf{(-19\%)} \\
\Xhline{1.2pt}

   & \textit{\textcolor{gray}{MBConv}} & \textbf{140.0} & \textbf{530.1} & \textbf{1865.3} & \textbf{3053.0} & \textbf{4114.5} & 1940.6 \\[-2pt]
  \textbf{28x28}  & \textit{\textbf{Gr}\textcolor{gray}{MBConv}} & 96.7 & 384.0 & 1350.1 & 2395.3 & 3769.6 & 1599.1 \footnotesize \textbf{(-17\%)} \\
\noalign{\setlength{\arrayrulewidth}{0.1pt}}
\cline{2-8}
\noalign{\setlength{\arrayrulewidth}{0.4pt}}
  (MMACs/ms)  & \textit{\textbf{Fu}\textcolor{gray}{MBConv}} & \textbf{752.8} & \textbf{3043.4} & \textbf{4882.2} & \textbf{5472.1} & 5676.0 & 3965.3 \\[-2pt]
    & \textit{\textbf{GrFu}\textcolor{gray}{MBConv}} & 395.0 & 1452.0 & 4237.8 & 4839.3 & \textbf{5703.9} & 3325.6 \footnotesize \textbf{(-16\%)} \\
\Xhline{1.2pt}

   & \textit{\textcolor{gray}{MBConv}} & \textbf{524.9} & \textbf{1503.4} & \textbf{2518.9} & \textbf{3820.4} & \textbf{4717.1} & 2617.0 \\[-2pt]
  \textbf{56x56}  & \textit{\textbf{Gr}\textcolor{gray}{MBConv}} & 379.8 & 1048.2 & 1953.9 & 3171.3 & 4416.1 & 2193.8 \footnotesize \textbf{(-16\%)} \\
\noalign{\setlength{\arrayrulewidth}{0.1pt}}
\cline{2-8}
\noalign{\setlength{\arrayrulewidth}{0.4pt}}
  (MMACs/ms)  & \textit{\textbf{Fu}\textcolor{gray}{MBConv}} & \textbf{2930.6} & \textbf{4971.6} & \textbf{6183.8} & \textbf{7226.3} & \textbf{7576.2} & 5777.7 \\[-2pt]
    & \textit{\textbf{GrFu}\textcolor{gray}{MBConv}} & 1402.0 & 3654.2 & 5068.1 & 6606.4 & 7288.1 & 4803.8 \footnotesize \textbf{(-17\%)} \\
\Xhline{1.2pt}

\Xhline{1.5pt}
\end{tabular}

\caption{Execution efficiency of MACs, measured in MMACs/ms on the Nvidia TITAN RTX GPU, for the four MBConv variants (MBConv, GrMBConv, FuMBConv and GrFuMBConv) for different channel dimensions and operating resolutions. Bold entries indicate whether the grouped or ungrouped variant has higher MACpS. All grouped variants have $groups=2$. It is similar to Table \ref{tab:mbconv_eff_exp}, but for resolutions $7\times7$, $14\times14$, $28\times28$ and $56\times56$ and only on GPU. }
\label{supp:tab:mbconv_eff_exp_gpu}
\end{table*}

\end{document}